\documentclass[10pt,twocolumn,letterpaper]{article}

\usepackage[pagenumbers]{iccv}

\usepackage{lineno}
\usepackage{adjustbox} 
\usepackage{array}
\usepackage{caption}
\usepackage{tabularx} 
\usepackage{ulem}
\usepackage[hyphens]{url}  
\newcommand\blfootnote[1]{%
  \begingroup
  \renewcommand\thefootnote{}\footnote{#1}%
  \addtocounter{footnote}{-1}%
  \endgroup
}
\definecolor{iccvblue}{rgb}{0.21,0.49,0.74}
\usepackage[pagebackref,breaklinks,colorlinks,allcolors=iccvblue]{hyperref}

\usepackage{graphicx}

\usepackage{booktabs}
\usepackage{multirow}
\usepackage{makecell}
\usepackage{colortbl}
\title{HATIR: Heat-Aware Diffusion for Turbulent Infrared Video Super-Resolution}

\author{
  Yang Zou\textsuperscript{\rm 1},
  Xingyue Zhu\textsuperscript{\rm 2},
  Kaiqi Han\textsuperscript{\rm 2},
  Jun Ma\textsuperscript{\rm 2},
  Xingyuan Li\textsuperscript{\rm 3},
  Zhiying Jiang\textsuperscript{\rm 4},
  Jinyuan Liu\textsuperscript{\rm 2}\footnotemark[2]\\
  \textsuperscript{\rm 1} Northwestern Polytechnical University, Xi'an, China \\
  \textsuperscript{\rm 2} Dalian University of Technology, Dalian, China \\
  \textsuperscript{\rm 3} Zhejiang University, Hangzhou, China \\
  \textsuperscript{\rm 4} Dalian Maritime University, Dalian, China \\
  {\tt\small atlantis918@hotmail.com}
}

\makeatletter
\let\@thanks\@empty   
\makeatother

\begin{document}
\twocolumn[{
\renewcommand\twocolumn[1][]{#1}%
\maketitle
\begin{center}
    \centering
    \captionsetup{type=figure}
    \includegraphics[width=0.8\textwidth]{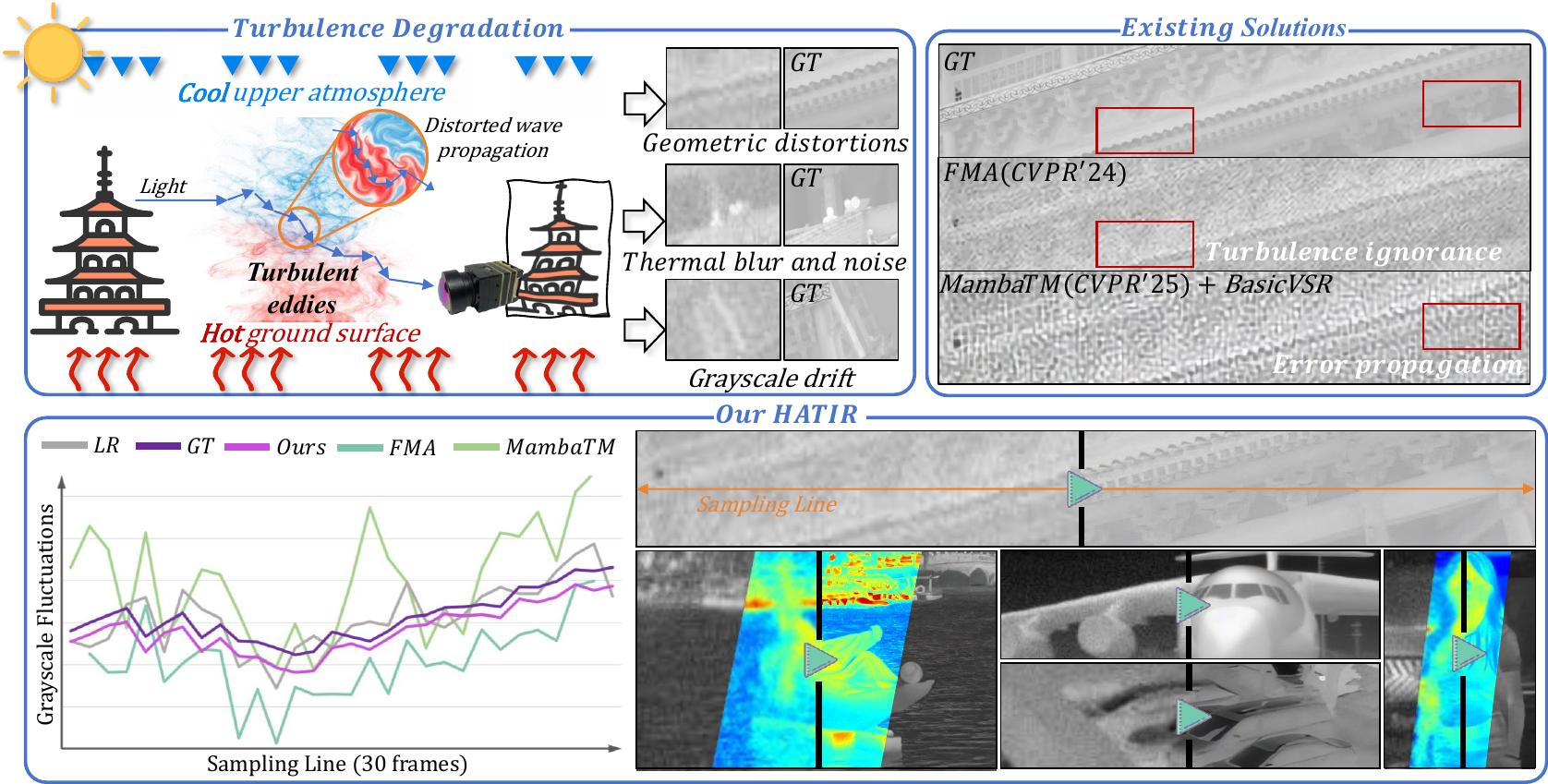}
    \captionof{figure}{Infrared VSR performance under turbulence conditions evaluated by HATIR on the proposed FLIR-IVSR dataset. The graph illustrates grayscale fluctuations along the orange-marked sampling line over time (30 video frames).}
    \label{fig:teaser}
\end{center}
}]

\begin{abstract}
\blfootnote{$^{\dagger}$ Corresponding author.}
Infrared video has been of great interest in visual tasks under challenging environments, but often suffers from severe atmospheric turbulence and compression degradation. Existing video super-resolution (VSR) methods either neglect the inherent modality gap between infrared and visible images or fail to restore turbulence-induced distortions. Directly cascading turbulence mitigation (TM) algorithms with VSR methods leads to error propagation and accumulation due to the decoupled modeling of degradation between turbulence and resolution. We introduce \textbf{HATIR}, a \textbf{H}eat-\textbf{A}ware Diffusion for \textbf{T}urbulent \textbf{I}nfra\textbf{R}ed Video Super-Resolution, which injects heat-aware deformation priors into the diffusion sampling path to jointly model the inverse process of turbulent degradation and structural detail loss. Specifically, HATIR constructs a Phasor-Guided Flow Estimator, rooted in the physical principle that thermally active regions exhibit consistent phasor responses over time, enabling reliable turbulence-aware flow to guide the reverse diffusion process. To ensure the fidelity of structural recovery under nonuniform distortions, a Turbulence-Aware Decoder is proposed to selectively suppress unstable temporal cues and enhance edge-aware feature aggregation via turbulence gating and structure-aware attention. We built FLIR-IVSR, the first dataset for turbulent infrared VSR, comprising paired LR-HR sequences from a FLIR T1050sc camera (\(1024 \times 768\)) spanning 640 diverse scenes with varying camera and object motion conditions. This encourages future research in infrared VSR. Project page: https://github.com/JZ0606/HATIR
\end{abstract}    
\section{Introduction}
High-quality infrared (IR) video is critical for vision tasks in challenging environments, such as autonomous driving, surveillance, and object tracking\cite{liu2025toward,wang2025efficient}. However, infrared imaging systems deployed in open atmospheric environments are highly susceptible to degradation caused by atmospheric turbulence. The formation of such turbulence is primarily attributed to the thermal and dynamic instability within the atmospheric boundary layer. Specifically, the temperature gradients between the hot ground surface and the cooler upper atmosphere generate convective flows that lead to the emergence of turbulent eddies across multiple spatial and temporal scales, as shown in Figure~\ref{fig:teaser}. These turbulent eddies cause random fluctuations of the refractive index and thermal radiation in the turbulence medium, which bend the propagated wave, resulting in geometric distortions, thermal blur, and grayscale drift in the infrared imaging~\cite{wang2023revelation,zou2024contourlet}. Compared to visible light cameras, IR sensors are more susceptible to turbulence-induced distortions due to their longer wavelengths and sensitivity to thermal fluctuations~\cite{liu2022target,li2024contourlet,li2025difiisr,liu2024promptfusion}. These real-world factors make the acquisition of high-quality IR video particularly challenging in practical scenarios. 

Conventionally, sliding-window based VSR methods~\cite{haris2019recurrent, tian2020tdan, wang2019edvr} reconstruct a high-resolution (HR) video by extracting features from a fixed number of adjacent frames within a short temporal window. Recurrent methods~\cite{huang2015bidirectional, huang2017video, isobe2020video, sajjadi2018frame} propagate hidden features by capturing long-term temporal dependencies and exploiting motion continuity across frames. Recently, diffusion-based methods~\cite{zhou2024upscale, yang2024motion, chen2024learning,Zhao_2023_ICCV_DDFM} have demonstrated remarkable performance in generating high-fidelity and perceptually realistic video content. These approaches primarily focus on incorporating temporal consistency strategies into the diffusion framework. 

Despite the remarkable progress of video super-resolution (VSR), existing approaches face two fundamental challenges when applied to infrared videos with turbulence: 1) \textbf{Modality gap.} Infrared images exhibit low texture contrast, weak structural boundaries, and thermal-dominated intensity patterns, deviating significantly from the assumptions underlying RGB-based VSR models~\cite{liu2022target,liu2025dcevo,zou2024enhancing,li2022drcr,li2025a2rnet,wang2025highlight}. 2) \textbf{Turbulence ignorance.} Severe atmospheric turbulence introduces nonlinear geometric distortions and unstable thermal boundaries, which are not explicitly addressed by conventional VSR pipelines. While turbulence mitigation (TM) methods fail to recover structural details. Simply cascading TM with VSR models often causes \textbf{error propagation and accumulation} due to their decoupled nature. Given these challenges, we ask, \textbf{“Is it possible to solve the turbulent infrared VSR through a unified inverse process?"}

The answer is \textbf{“Yes."} We propose HATIR, a Heat-Aware Diffusion framework for Turbulent InfraRed Video Super-Resolution, which injects physically grounded heat-aware deformation priors into the diffusion sampling path to jointly model the inverse process of turbulence degradation and structural detail loss. By unifying alignment and restoration in a single generative path, HATIR mitigates error amplification caused by misalignment and thermal blur, which conventional approaches often struggle with. Specifically, we propose Phasor-Guided Flow Estimator (PhasorFlow), enabling robust turbulence-aware motion guidance. Also, a Turbulence-Aware Decoder (TAD) is introduced to enhance structural fidelity under non-uniform distortions via turbulence-aware gating and structure-aware feature fusion. To benchmark this task, we construct the first dataset for turbulent infrared VSR, enabling evaluation under long-range infrared degradation. Our contribution can be summarized as follows:

\begin{itemize}
\item We introduce \textbf{HATIR}, a \textbf{H}eat-\textbf{A}ware Diffusion for \textbf{T}urbulent \textbf{I}nfra\textbf{R}ed Video Super-Resolution, which jointly models the degradation process of turbulent degradation and structural detail loss through physics-driven heat-aware deformation priors.

\item We design a phasor-guided flow estimator, rooted in thermal consistency, to provide robust turbulence-aware guidance for reverse diffusion. A Turbulence-Aware Decoder is further introduced to enhance structural restoration by suppressing unstable temporal information and reinforcing edge-aware feature aggregation.

\item We built the first dataset for the turbulent infrared VSR task, FLIR-IVSR, comprising paired LR-HR sequences captured by a FLIR T1050sc camera at a resolution of 1024 \(\times\) 768. FLIR-IVSR spans 640 diverse scenes under varying camera and object motion conditions.
\end{itemize}

\section{Related Work}
\subsection{Video Super-Resolution}
Existing VSR methods can be broadly categorized into multiple-input single-output (MISO) and multiple-input multiple-output (MIMO) paradigms. MISO-based methods reconstruct the center frame from a fixed window of LR frames. This line of work includes filter-based approaches~\cite{jo2018deep}, alignment-based methods using deformable convolutions~\cite{wang2019edvr}, and attention-based designs~\cite{li2020mucan}. Recent extensions further integrate motion-aware modules~\cite{youk2024fma}, recurrent propagation~\cite{chi2024egovsr}, or G-buffer priors~\cite{zheng2025efficient} for enhanced temporal modeling and efficiency. MIMO-based methods jointly reconstruct multiple frames, allowing for consistent modeling across time. This includes transformer-based architectures~\cite{liang2022recurrent} and diffusion-driven approaches~\cite{yang2024motion,zhou2024upscale}, which incorporate motion priors into the generative process to improve fidelity and coherence.

\subsection{Video Turbulence Mitigation}
Traditional methods typically employ a three-stage pipeline comprising registration, fusion, and deblurring. Recent learning-based methods address turbulence dynamics in an end-to-end manner. DATUM~\cite{zhang2024spatio} decouples alignment and content restoration across short sequences. MambaTM~\cite{zhang2025learning} adopts state space models for efficient long-range temporal modeling. Turb-Seg-Res~\cite{saha2024turb} separates motion-dominant regions for region-specific refinement. Nevertheless, these methods are designed for RGB videos and struggle in infrared domains due to weak textures and thermal blur. Moreover, they typically address turbulence alone, overlooking the resolution degradation that coexists in real infrared settings. This highlights the need for a unified solution to jointly mitigate turbulence and enhance resolution in infrared videos.

\begin{figure*}[!h]
    \centering
    \includegraphics[width=1\linewidth]{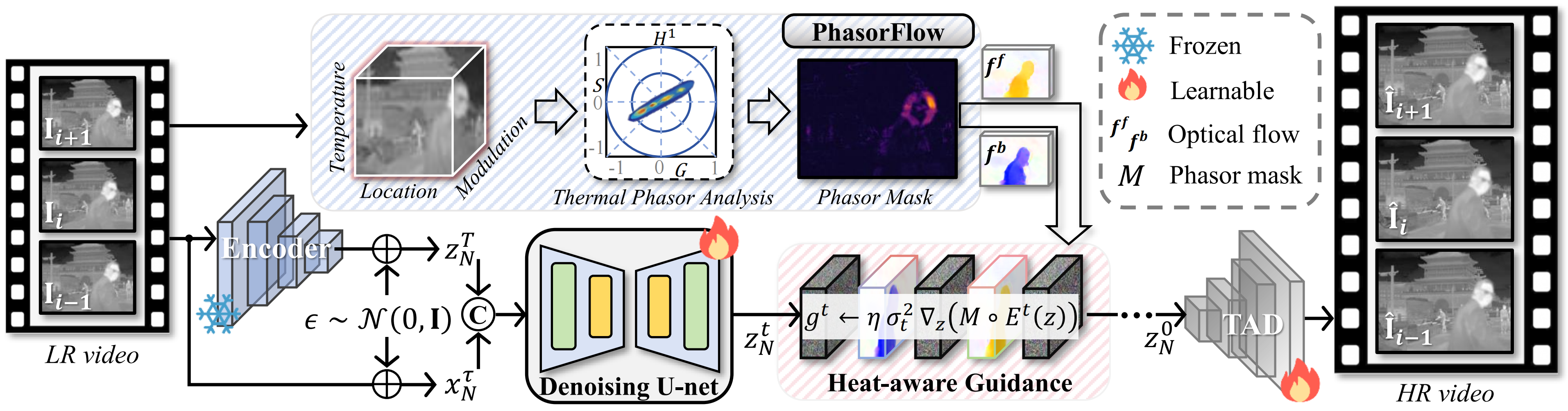}
    \caption{Given a low-resolution (LR) turbulent infrared video sequence $\mathbf{I}_{LR} = \{\mathbf{I}_1, \mathbf{I}_2, \dots, \mathbf{I}_N\}$, HATIR reconstructs a high-resolution (HR) sequence $\mathbf{I}_{HR} = \{\hat{\mathbf{I}}_1, \hat{\mathbf{I}}_2, \dots, \hat{\mathbf{I}}_N\}$ with suppressed turbulence distortions and enhanced temporal coherence. The proposed unified latent diffusion framework jointly addresses spatial degradation removal and inter-frame alignment for infrared videos under atmospheric turbulence.}
    \label{fig:pipeline}
\end{figure*}

\section{Method}
\subsection{Overview}
As illustrated in Figure~\ref{fig:pipeline}, the LR video is first encoded into a latent space via a VAE encoder. Then, guided by the proposed PhasorFlow, which captures the thermal dynamics of time-varying heat sources, the diffusion model iteratively refines the latent variables under turbulence-aware modulation. Finally, a Turbulence-Aware Decoder (TAD) reconstructs the HR frames by suppressing unreliable temporal cues and reinforcing edge structures.

\subsection{Phasor-Guided Flow Estimator}
\begin{figure}
    \centering
    \includegraphics[width=1\linewidth]{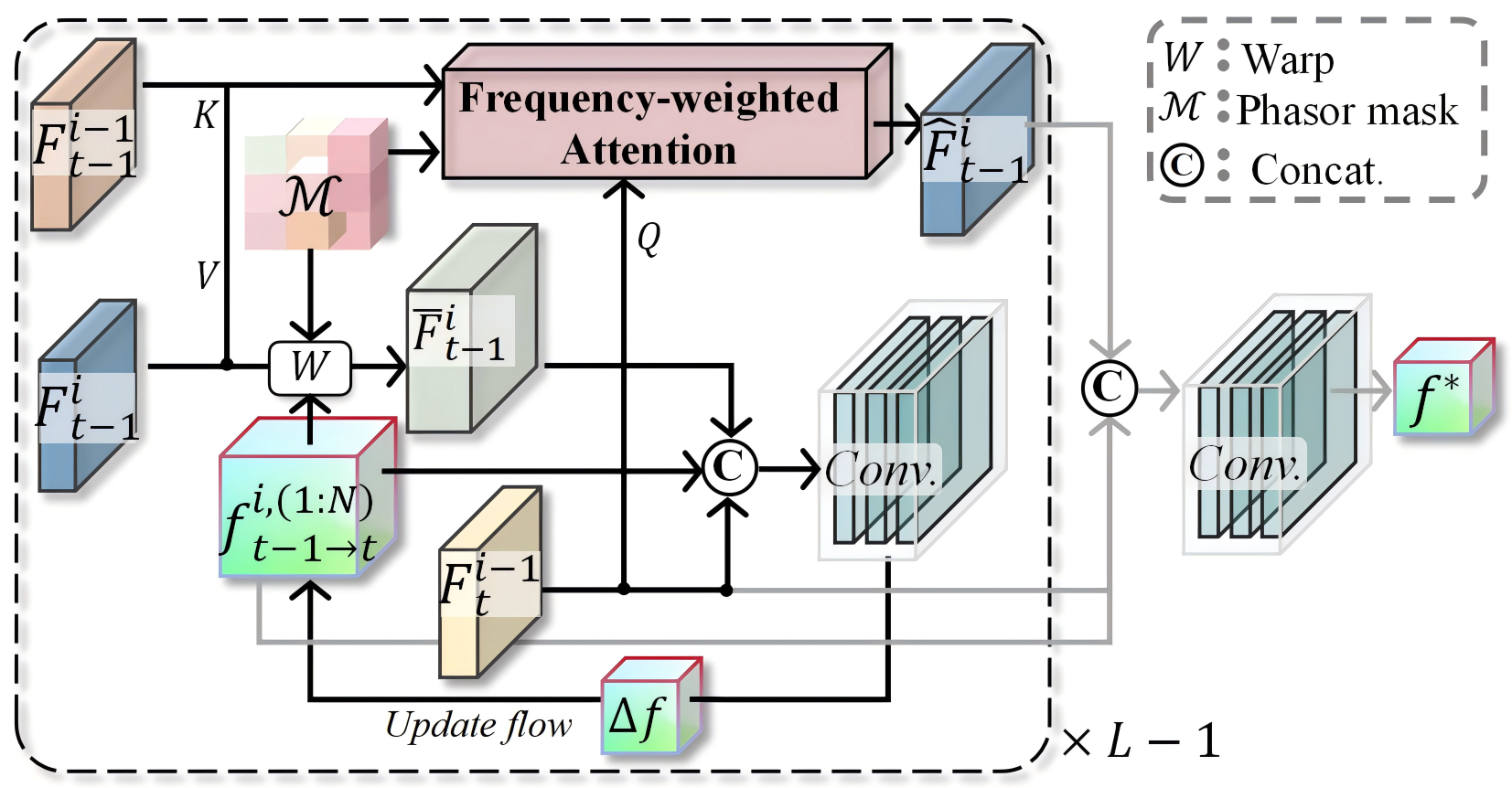}
    \caption{Overview of PhasorFlow.}
    \label{fig:phasor_flow}
\end{figure}

To tackle turbulence-induced distortions and detail degradation in low-resolution infrared videos, we propose Phasor-Guided Flow Estimator (PhasorFlow), a heat-aware flow estimator that guides diffusion sampling with thermal priors as shown in Figure~\ref{fig:phasor_flow}. While prior works leverage optical flow for inter-frame alignment~\cite{yang2024motion, liang2022recurrent, zhou2024upscale,10516464}, they often fail in turbulent infrared settings due to weak textures, ambiguous boundaries, and the stochastic nature of turbulence. PhasorFlow addresses these issues by introducing Frequency-Weighted Attention, guided by thermal phasor analysis, which measures the temporal consistency of thermal radiation in the frequency domain.

Specifically, we first extract shallow features $F^0 \in \mathbb{R}^{T \times H \times W \times C}$ and segment them into short clips. For each clip $F^i_t$, an initial flow $f^i_{t-1 \rightarrow t}$ is estimated via a pretrained flow network~\cite{ranjan2017optical}, and iteratively refined using the Phasor Mask and Frequency-Weighted Attention in a locally parallel, globally recurrent manner.

\subsubsection{Phasor Mask}
To robustly identify thermally stable regions under turbulence, we calculate the Phasor Mask to assess the temporal frequency response of infrared sequences. This is based on the physical observation that heat-emitting regions exhibit stable temporal dynamics, while turbulence causes high-frequency, spatially varying perturbations.

Given a short infrared sequence \(\mathbf{I} \in \mathbb{R}^{B \times T \times 1 \times H \times W}\), we first reshape it to \(\mathbf{I}' \in \mathbb{C}^{B \times H \times W \times T}\) and compute the discrete Fourier transform (DFT) over the temporal dimension as \(\hat{\mathbf{I}}(x) = \mathcal{F}_t \left( \mathbf{I}(x, :) \right), \quad x \in \Omega\). We then extract the magnitude of the first harmonic (e.g., \(\hat{\mathbf{I}}_1(x)\)) as the primary frequency response by \(M_{\text{phasor}}(x) = \left| \hat{\mathbf{I}}_1(x) \right|\). Finally, \(M_{\text{phasor}}\) is normalized to [0,1] to serve as a soft mask:
\begin{equation}
M_{\text{phasor}}(x) = \sigma \left( \alpha \cdot \left( M_{\text{phasor}}(x) - \mu \right) \right),
\end{equation}

\noindent where \(\mu\) is the spatial mean and \(\alpha\) is a scaling factor. This Phasor Mask emphasizes pixels with consistent temporal thermal signatures and is integrated into attention modulation and flow guidance to suppress unstable turbulent regions and preserve heat-sensitive structural information.

\subsubsection{Frequency-weighted Attention}
Given the $(t{-}1)$-th clip feature $F_{t-1}^i$ from the $i$-th layer, our objective is to estimate the turbulence-mitigated flow $\hat{f}_{t-1\rightarrow t}^{i,(1:N)}$ across the $N$ frames in each clip. For each flow $\hat{f}_{t-1\rightarrow t, n'}^{i,(n)}$ aligning frame $n'$ in clip $t{-}1$ to frame $n$ in clip $t$, we first compute a coarse optical flow $f_{t-1\rightarrow t}^{i,(1:N)}$ using SpyNet~\cite{ranjan2017optical}, and obtain coarse aligned features via:
\begin{equation}
\bar{F}_{t-1}^{i,(1:N)} = \text{Warp}(F_{t-1}^i,\ M_{phasor,t-1\rightarrow t}^{(1:N)} \circ f_{t-1\rightarrow t}^{i,(1:N)}),
\end{equation}

\noindent where $M_{phasor}$ denotes the thermal stability prior from Phasor Mask. These coarse features are concatenated with the current frame and flow to predict flow residuals via a CNN:
\begin{equation}
\Delta f_{t-1\rightarrow t}^{i,(1:N)} = \text{Conv}(\text{Concat}(\bar{F}_{t-1}^{i,(1:N)}, F_t^{i-1}, f_{t-1\rightarrow t}^{i,(1:N)})).
\end{equation}

We then update the flow through an averaged refinement across $M$ predicted offsets:
\begin{equation}
      f_{t-1\rightarrow t, n'}^{i+1,(n)} = f_{t-1\rightarrow t, n'}^{i,(n)} + \frac{1}{M}\sum_{m=1}^M\{\Delta f_{t-1\rightarrow t,n'}^{i,(n)}\}_m,
\end{equation}

\noindent where $\{\Delta f_{t-1\rightarrow t,n'}^{i,(n)}\}_m$ denotes the $m$-th offset in total $M$ predictions.

To enhance feature reliability during turbulence, we sample features via the updated flow and apply phasor-guided attention. Specifically, the attention queries, keys, and values are defined as \(Q = F_{t,n}^{i-1} P_Q\), \(K = \text{Sampling}(F_{t-1}^{i-1} P_K,\ f + \Delta f)\), and \(V = \text{Sampling}(F_{t-1}^{i} P_V,\ f + \Delta f)\), where $f + \Delta f$ denotes the total motion offset. The Phasor Mask modulates attention weights as:
\begin{equation}
\hat{F}_{t-1}^{i,(n)} = (M_{phasor}^{(n)} \circ \mathcal{S}(QK^\top / \sqrt{C})) V + \text{MLP}(\hat{F}_{t-1}^{i,(n)}),
\end{equation}

\noindent where \(\mathcal{S}\) denotes the \(\text{SoftMax}\) operation. In the final layer $L$, we recompute the offset using the refined feature \(\hat{F}_{t-1}^{L,(1:N)}\) to update the final flow:
\begin{equation}
\scriptsize
f_{t-1 \rightarrow t,n'}^{\ast} = f + 
\frac{1}{M} \sum_{m=1}^{M} 
\bigg[ \overbrace{ \mathcal{H} \left( \hat{F}_{t-1}^{L,(1:N)},\ F_t^{L-1},\ f_{t-1 \rightarrow t}^{L,(1:N)} \right) }^{\Delta f_{t-1\rightarrow t}^{L,(1:N)}} \bigg]_{n'}^{(m)},
\end{equation}

\noindent where \(f\) represents \(f_{t-1 \rightarrow t,n'}^{L}\), \(\mathcal{H}(\cdot)\) denotes a lightweight convolutional network.

\subsubsection{Heat-aware Guidance}
To improve the stability and consistency of the denoising trajectory under turbulence, we inject a physics-informed guidance term derived from thermal motion priors. At each denoising step $t$, we first define the symmetric warping error between bidirectional flows:
\begin{equation}
\small
\begin{aligned}
    E^t(z) = &\sum_{i=1}^{N-1}\|(\text{Warp}(z^t_i,f_{b,i}^*)-z^t_{i+1}\|_1 \\ + &\sum_{i=2}^{N}\|(\text{Warp}(z^t_i,f_{f,i-1}^*)-z^t_{i-1}\|_1,
\end{aligned}
\end{equation}

\noindent where $f_{f,i-1}^*$ and $f_{b,i}^*$ are the forward and backward flows estimated by PhasorFlow. To localize reliable temporal structures, we construct a heat-aware modulation mask $M_{\text{joint}}$ by fusing an occlusion-aware mask and the normalized thermal Phasor Mask as \(M_{\text{joint}} = M_{\text{occ}} \cdot M_{\text{phasor}},\) where $M_{\text{phasor}}$ denotes the Phasor Mask. 

The final heat-aware guidance term is defined as \(g^t = \eta\, \sigma_t^2\, \nabla_z \left( M_{\text{joint}} \circ E^t(z) \right)\), where $\sigma_t^2$ is the noise variance at step $t$, and $\eta$ modulates the influence of the guidance. The denoising step is then adjusted as:
\begin{equation}
\hat{z}^t = z^{t+1} - \sigma_t^2 \epsilon_\phi(z^{t+1}, t) - g^t,
\end{equation}

\noindent where $\epsilon_\phi$ denotes the noise prediction network of the diffusion model. This guidance steers the sampling trajectory toward temporally coherent and thermally stable representations, which are subsequently decoded by the Turbulence-Aware Decoder (TAD).

\subsection{Turbulence-Aware Decoder}
IR images typically exhibit weak textures, blurred thermal boundaries, and reduced structural saliency compared to visible images. These properties, compounded by atmospheric turbulence, result in alignment errors and unreliable motion estimation. Also, enforcing strict temporal consistency in turbulence-distorted regions may introduce erroneous corrections. Given those issues, we propose the Turbulence-Aware Decoder (TAD) to enhance temporal coherence while selectively mitigating turbulence-induced distortions.

\subsubsection{Turbulence Mask Gating}
Given the latent feature $z_t$ at time step $t$, we first apply temporal convolutions to extract inter-frame dependencies. To identify turbulence-corrupted regions, we construct a disturbance heatmap $T_{\text{map}}$ based on bidirectional warping errors:
\begin{equation}
\begin{aligned}
    T_{\text{map}} = & \left\| \text{Warp}(x_{t-1}, f_{t \to t-1}) - x_t \right\|_1 +  \\
    & \left\| \text{Warp}(x_{t+1}, f_{t \to t+1}) - x_t \right\|_1,
\end{aligned}
\end{equation}

\noindent where $f_{t \rightarrow t\pm1}$ denotes bidirectional optical flows estimated by the PhasorFlow module. The heatmap is converted to a gating mask $G \in [0,1]^{H \times W}$ via \(G = \sigma\left( \text{Conv}_{1\times1}(T_{\text{map}}) \right)\), which modulates the temporal convolution output in a residual manner as:
\begin{equation}
f_t = \text{TMG}(z_t) = G \circ \text{Conv}_{1\times1}(\text{ResBlock}(z_t)).
\end{equation}

This mechanism adaptively filters out turbulence-corrupted regions, ensuring that cross-frame modeling is restricted to structurally stable areas.

\subsubsection{IR Structure-Aware Attention}
To further enhance the temporal alignment of critical structures, we introduce IR-SAA, which selectively enforces consistency in high-frequency regions (e.g., edges, contours) while avoiding redundant alignment in low-saliency regions.

From the output $f_t$ of TMG, we construct a structure attention map $A_t \in [0,1]^{H \times W}$ using the gradient magnitude as \(A_t = \sigma\left( \text{Conv}_{1\times1}\left( \left\| \nabla f_t \right\|_1 \right) \right)\), and enhance the feature via residual attention \(f^{\text{enh}}_t = f_t + \lambda (f_t \circ A_t)\), where $\lambda$ is a fixed scaling coefficient, this allows the model to focus computational capacity on thermally relevant structures.

\begin{figure*}[!h]
    \centering
    \includegraphics[width=\textwidth]{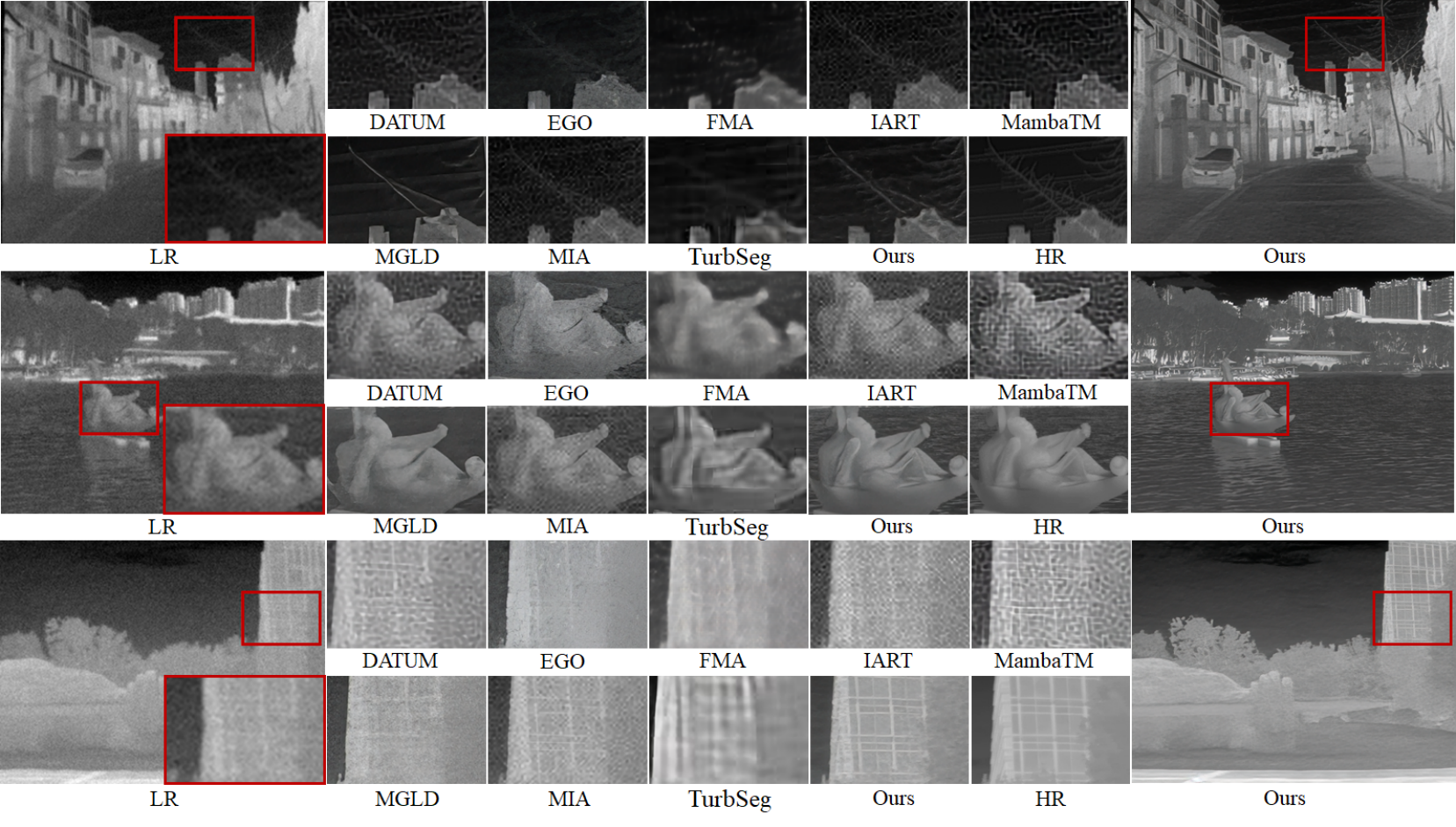}
    \caption{Qualitative results. The first row is from the static scenes of the $\mathrm{M^3FD}$ dataset, while the second and third rows are from the FLIR-IVSR dataset. MambaTM, DATUM, and Turb-Seg are combined with BasicVSR to form a two-stage pipeline.}
    \label{fig:turbulence-comparison}
\end{figure*}

\subsubsection{Optimization}
We fine-tune the TAD on top of a pre-trained VAE decoder for turbulent infrared VSR tasks. We first define the Thermal Reconstruction Loss to emphasize high-fidelity recovery in thermally active regions as \(\mathcal{L}_{\text{thermal}} = \left\| (\hat{\mathbf{I}} - \mathbf{I}_{\text{gt}}) \circ M_{\text{phasor}} \right\|_1\), where $M_{\text{phasor}}$ is Phaser Mask from thermal phasor analysis. To encourage sharper recovery of blurred thermal contours, we introduce the Thermal Edge Loss as \(\mathcal{L}_{\text{edge}} = \left\| (\nabla \hat{\mathbf{I}} - \nabla \mathbf{I}_{\text{gt}}) \circ M_{\text{phasor}} \right\|_1\), where $\nabla(\cdot)$ denotes a Laplacian operator applied for edge extraction to penalizes misalignment in thermal edge structures. Also, to preserve temporal consistency across the reconstructed sequence, we employ a Frame Difference Loss defined as \(\mathcal{L}_{\text{diff}} = \sum_i \left\| (\hat{\mathbf{I}}_{i+1} - \hat{\mathbf{I}}_i) - (\mathbf{I}^{\text{gt}}_{i+1} - \mathbf{I}^{\text{gt}}_i) \right\|_1\). 

The total loss function is then formulated by combining those loss functions. This joint loss not only enhances restoration in thermal-sensitive regions but also improves stability of the overall diffusion trajectory under turbulence.

\section{Experiments}
\subsection{Experimental Settings}
\subsubsection{Implementation Details}
Our network is trained on an NVIDIA A800 GPU using the Adam optimizer, with hyperparameters set to $\beta_1$ = 0.9 and $\beta_2$ = 0.999. We first fine-tune the U-Net backbone, initializing it with pretrained weights from Stable Diffusion v2.1~\cite{rombach2022high}. To effectively incorporate information from LR inputs, we introduce a lightweight time-aware encoder that extracts temporal features from LR images and encodes them as conditional inputs to guide the diffusion process. Subsequently, we train the proposed PhasorFlow module independently and integrate it with the fine-tuned U-Net to perform image sampling, which generates latent features for training the Turbulence-Aware Decoder. 

\begin{table*}[!ht]
\centering
\small
\renewcommand{\arraystretch}{1.2}
\resizebox{\textwidth}{!}{
\begin{tabular}{@{\hspace{5pt}}lccccccccccccccc@{\hspace{5pt}}}
\Xhline{1.1pt}

\multicolumn{1}{c}{\textbf{Datasets}} 
& \multicolumn{5}{c}{\textbf{Set5}} 
& \multicolumn{5}{c}{\textbf{Set10}} 
& \multicolumn{5}{c}{\textbf{Set20}} \\
\cmidrule(lr){1-1} \cmidrule(lr){2-6} \cmidrule(lr){7-11} \cmidrule(lr){12-16}
 
\multicolumn{1}{c}{\textbf{Methods}}
& PSNR$\uparrow$ & SSIM$\uparrow$ & LPIPS$\downarrow$ & DISTS$\downarrow$ & VMAF$\uparrow$ 
& PSNR$\uparrow$ & SSIM$\uparrow$ & LPIPS$\downarrow$ & DISTS$\downarrow$ & VMAF$\uparrow$ 
& PSNR$\uparrow$ & SSIM$\uparrow$ & LPIPS$\downarrow$ & DISTS$\downarrow$ & VMAF$\uparrow$ \\
\hline

\multicolumn{16}{c}{\textbf{\boldmath$\mathrm{M^{3}FD}$}} \\
\hline
MambaTM   & 25.6757 & 0.5741 & 0.4078 & 0.2319 & 28.0380  
          & 25.6237 & 0.5822 & 0.4084 & 0.2231 & 26.7815  
          & 25.6078 & 0.5779 & 0.4095 & 0.2299 & 26.3855 \\
Turb-seg  & 22.1135 & 0.6857 & 0.2976 & 0.2402 & 5.7285  
          & 24.2823 & 0.7399 & 0.2582 & 0.2084 & 8.4070  
          & 23.7615 & 0.7361 & 0.2584 & 0.2152 & 6.1068 \\
DATUM     & \underline{28.1310} & 0.6749 & 0.3569 & 0.1880 & \textbf{45.8987}  
          & \underline{28.3336} & 0.7020 & 0.3420 & 0.1771 & \underline{45.6202}  
          & \underline{28.4232} & 0.7026 & 0.3389 & 0.1807 & \underline{46.3185} \\
MGLDVSR   & 27.1603 & \underline{0.8003} & \underline{0.2106} & \textbf{0.1513} & 26.7114  
          & 27.9049 & \underline{0.7965} & \underline{0.1919} & \underline{0.1612} & 25.5742  
          & 28.1681 & \underline{0.8137} & \underline{0.1902} & \textbf{0.1515} & 27.5137 \\
FMA-NET   & 27.5482 & 0.7831 & 0.2376 & 0.2200 & 31.3568  
          & 27.0874 & 0.7784 & 0.2344 & 0.2105 & 29.4062  
          & 27.1545 & 0.7788 & 0.2324 & 0.2139 & 28.9050 \\
MIA-VSR   & 27.7264 & 0.6153 & 0.3529 & 0.2461 & 44.9468  
          & 27.6816 & 0.6240 & 0.3576 & 0.2533 & 45.1764  
          & 27.7188 & 0.6221 & 0.3599 & 0.2534 & 44.9845 \\
IART      & 27.7020 & 0.6020 & 0.3528 & 0.2542 & \underline{45.6605}  
          & 27.6319 & 0.6114 & 0.3576 & 0.2607 & 45.5397  
          & 27.6641 & 0.6089 & 0.3605 & 0.2608 & 45.3884 \\
EGOVSR    & 26.6591 & 0.7230 & 0.2611 & 0.1975 & 27.1438  
          & 26.2876 & 0.7055 & 0.2865 & 0.1971 & 25.0401  
          & 26.3767 & 0.7102 & 0.2833 & 0.1992 & 24.9732 \\
\rowcolor[gray]{0.9}
Ours      & \textbf{29.7819} & \textbf{0.8311} & \textbf{0.1724} & \underline{0.1576} & 44.6731  
          & \textbf{30.6093} & \textbf{0.8352} & \textbf{0.1455} & \textbf{0.1479} & \textbf{48.6273}  
          & \textbf{30.3834} & \textbf{0.8370} & \textbf{0.1555} & \underline{0.1530} & \textbf{46.6925} \\
\hline
\multicolumn{16}{c}{\textbf{FLIR-IVSR}} \\ 
\hline
MambaTM   & 22.7972 & 0.3114 & 0.6511 & 0.3369 & 11.6541  & 23.3786 & 0.3256 & 0.6693 & 0.3654 & 12.8628  & 23.7571 & 0.3665 & 0.6267 & 0.3399 & 18.0775 \\
Turb-seg  & 24.8976 & \underline{0.7509} & \underline{0.2973} & 0.2346 & 4.6408   & 23.0295 & \underline{0.7825} & \underline{0.2770} & 0.2559 & 4.3775   & 20.2981 & 0.6894 & 0.6375 & 0.3606 & 4.3461 \\
DATUM     & 27.6349 & 0.5596 & 0.5063 & 0.2674 & 25.9121  & 27.9964 & 0.5688 & 0.5230 & 0.2981 & 24.4874  & 27.1081 & 0.5550 & 0.5156 & 0.2831 & 29.4297 \\
MGLDVSR   & 29.2938 & 0.6336 & 0.3679 & \underline{0.2274} & 28.1148  & \underline{30.4112} & 0.7045 & 0.3519 & \underline{0.2476} & 27.9592  & 27.5376 & \underline{0.7983} & \underline{0.2072} & \underline{0.1608} & 25.5895 \\
FMA-NET   & \underline{29.6184} & 0.7457 & 0.3177 & 0.2618 & 28.6511  & 29.5584 & 0.7773 & 0.2843 & 0.2741 & 23.4312  & 28.0662 & 0.7261 & 0.3244 & 0.2710 & 26.2214 \\
MIA-VSR   & 27.2881 & 0.4882 & 0.5169 & 0.3515 & 25.8345  & 27.5797 & 0.4920 & 0.5244 & 0.3752 & 24.1688  & 27.1045 & 0.4927 & 0.5171 & 0.3625 & 29.9711 \\
IART      & 27.2596 & 0.4893 & 0.5008 & 0.3573 & 26.4507  & 27.5574 & 0.4940 & 0.5018 & 0.3815 & 24.7818  & 27.0212 & 0.4877 & 0.5024 & 0.3697 & 30.1541 \\
EGOVSR    & 28.7845 & 0.6452 & 0.3835 & 0.2629 & \underline{36.7992}  & 29.5250 & 0.6645 & 0.4177 & 0.2938 & \underline{35.5069}  & \underline{28.4134} & 0.6573 & 0.3873 & 0.2733 & \underline{32.1390} \\
\rowcolor[gray]{0.9}
Ours      & \textbf{33.3719} & \textbf{0.8683} & \textbf{0.1227} & \textbf{0.1183} & \textbf{46.6922} 
          & \textbf{33.8680} & \textbf{0.8545} & \textbf{0.1555} & \textbf{0.1559} & \textbf{42.8454} 
          & \textbf{32.4682} & \textbf{0.8415} & \textbf{0.1377} & \textbf{0.1464} & \textbf{44.8895} \\
\Xhline{1.1pt}
\end{tabular}
}
\caption{Quantitative comparison on \(\mathrm{M^{3}FD}\) and FLIR-IVSR. The best is in \textbf{bold}, while the second is \underline{underlined}. For $\mathrm{M^{3}FD}$, Set5/10/20 are randomly sampled subsets. For FLIR-IVSR, the three sets correspond to “camera-static (static scene)", “camera-static (dynamic scene)", and “camera-moving", respectively.}

\label{table:final_metrics_layout}
\end{table*}

\subsubsection{Datasets and Evaluation Metrics}
To facilitate research in infrared video super-resolution under atmospheric turbulence, we construct FLIR-IVSR, an infrared VSR dataset comprising 640 paired LR-HR infrared video sequences captured using a FLIR T1050sc thermal camera at a resolution of \(1024 \times 768\). The dataset encompasses a wide range of motion patterns and scene categories, and is divided into two subsets based on camera motion. The camera-moving subset contains 135 sequences, featuring scenarios with platform-induced motion. The camera-static subset includes 510 sequences, further categorized into: (i) Dynamic scenes (495 sequences), characterized by object-level or environmental motion with a stationary camera; (ii) Static scenes (15 sequences) with minimal motion. FLIR-IVSR provides a comprehensive and challenging benchmark for assessing infrared VSR methods under severe low-resolution and turbulence-induced degradations. The process of building the FLIR-IVSR is detailed in the supplementary materials.

We train all models on the FLIR-IVSR training set, which consists of 505 turbulent infrared video sequence LR-HR pairs. Evaluation is conducted on two test sets: (1) the FLIR-IVSR test set comprising 135 turbulent infrared video pairs, and (2) a synthetic turbulence benchmark constructed from the static scenes of the public $\mathrm{M^3FD}$ dataset by simulating turbulence-induced distortions.

To comprehensively assess both fidelity and perceptual quality, we report five widely used metrics: Peak Signal-to-Noise Ratio (PSNR), Structural Similarity Index Measure (SSIM), Learned Perceptual Image Patch Similarity (LPIPS), Deep Image Structure and Texture Similarity (DISTS), and Video Multi-Method Assessment Fusion (VMAF). Detailed definitions of these metrics are provided in~\cite{ma2019infrared}.

\subsubsection{Comparative Methods}
We perform a comprehensive comparison of our approach with five video super-resolution(VSR) methods, including MIA-VSR~\cite{zhou2024video}, FMA-Net~\cite{youk2024fma}, EGOVSR~\cite{chi2024egovsr}, IART~\cite{xu2024enhancing}, and MGLDVSR~\cite{yang2024motion}, as well as three turbulence removal methods, MambaTM~\cite{zhang2025learning}, DATUM~\cite{zhang2024spatio}, and Turb-Seg~\cite{saha2024turb}. Notably, each turbulence removal method is combined with a unified VSR model, BasicVSR~\cite{chan2022basicvsr}, forming two-stage pipelines that perform turbulence correction followed by resolution enhancement.

\subsection{Qualitative Results}
To visually demonstrate the effectiveness of our method, Figure \ref{fig:turbulence-comparison} presents the restoration results of different approaches on the same frame of identical samples. The top sample comes from the turbulence-degraded $\mathrm{M^{3}FD}$ dataset, while the bottom two are from our FLIR-IVSR dataset. As observed in the figure, the three two-stage approaches— MambaTM, DATUM, and Turb-Seg—that perform turbulence mitigation followed by super-resolution suffer from error accumulation during turbulence removal, and their subsequent super-resolution steps further amplify these artifacts. The four VSR methods—MIA-VSR, FMA-Net, EGOVSR, and IART— also fail to effectively address the noise, blurring, and spatial distortion caused by turbulence. While the diffusion-based VSR method MGLDVSR shows some capability in recovering blurred and noisy content, it still exhibits texture loss and restoration errors due to the lack of turbulence mitigation and infrared-specific guidance. In contrast, our approach successfully restores thermal details and spatial distortions while preserving high-resolution texture and maintaining the visual characteristics intrinsic to infrared imagery.

\begin{table}[!tp]
\centering
\footnotesize
\resizebox{\linewidth}{!}{%
\begin{tabular}{lccccc}
\Xhline{1.1pt}
 Guide & PSNR $\uparrow$ & SSIM $\uparrow$ & LPIPS $\downarrow$ & DISTS $\downarrow$ & VMAF $\uparrow$ \\
\hline
 PhasorFlow & \textbf{33.6507} & \textbf{0.8535} & \textbf{0.1377} & \textbf{0.1482} & \textbf{45.3972} \\
SpyNet     & 28.9387 & 0.7668 & 0.2386 & 0.1615 & 33.1466 \\
\Xhline{1.1pt}
\end{tabular}
}
\caption{Quantitative ablation study on PhasorFlow.}

\label{tab:flow_ablation}
\end{table}

\begin{figure}[!tp]
    \centering
    \includegraphics[width=0.47\textwidth]{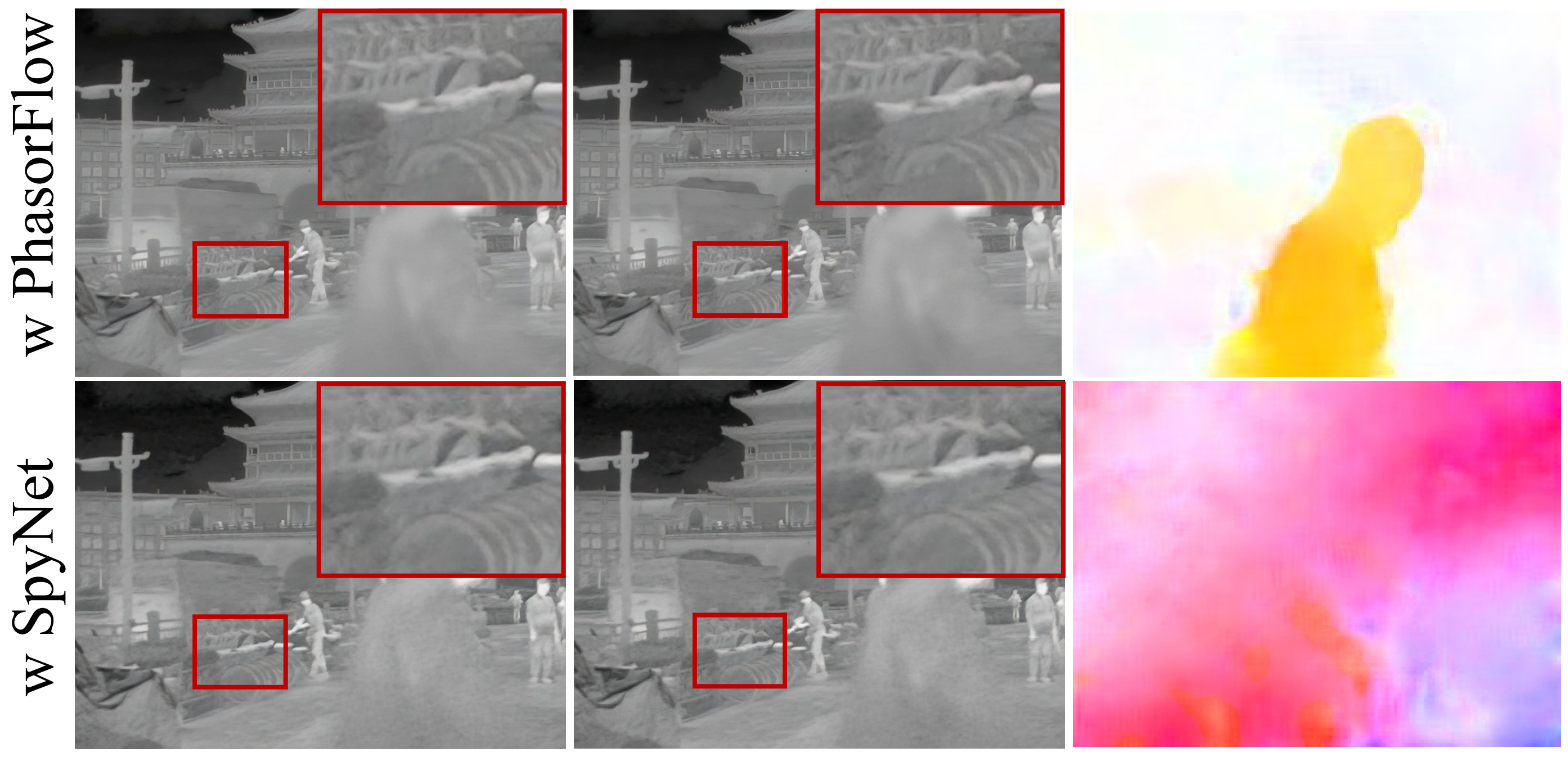}
    \caption{Qualitative ablation on the PhasorFlow.}
    \label{fig:ablation_phasorflow}
\end{figure}

\subsection{Quantitative Comparison}
Table~\ref{table:final_metrics_layout} compares the quantitative results on the FLIR-IVSR and turbulence-degraded $\mathrm{M^{3}FD}$ datasets. For $\mathrm{M^{3}FD}$, subsets are randomly sampled for evaluation, while for FLIR-IVSR, test samples are selected from different scene categories to enable a comprehensive analysis. As demonstrated in the table, our method achieves the best performance among all compared approaches across different camera motions on the FLIR-IVSR dataset. For the $\mathrm{M^{3}FD}$ dataset, we show clear advantages on the larger test sets (sizes 10 and 20), demonstrating the effectiveness of our method in mitigating complex degradation conditions for infrared VSR.

\begin{table}[!tp]
\centering
\footnotesize
\resizebox{\linewidth}{!}{%
\begin{tabular}{@{\hspace{2pt}}ccccccc@{\hspace{2pt}}}
\Xhline{1.1pt}
IR--SAA & TMG & PSNR$\uparrow$ & SSIM$\uparrow$ & LPIPS$\downarrow$ & DISTS$\downarrow$ & VMAF$\uparrow$ \\
\hline
- & - & 26.3283 & 0.6775 & 0.2862 & 0.1987 & 32.0941 \\
\checkmark & - & 27.3985 & 0.7169 & 0.1735 & 0.1735 & 36.2274 \\
- & \checkmark & 28.4125 & 0.7418 & 0.1564 & 0.1541 & 40.9598 \\
\checkmark & \checkmark & \textbf{32.2391} & \textbf{0.8229} & \textbf{0.1358} & \textbf{0.1431} & \textbf{43.4152} \\
\Xhline{1.1pt}
\end{tabular}
}
\caption{Quantitative ablation on the TAD.}
\label{tab:ablation_tad}
\end{table}

\begin{figure}[!tp]
    \centering
    \includegraphics[width=0.47\textwidth]{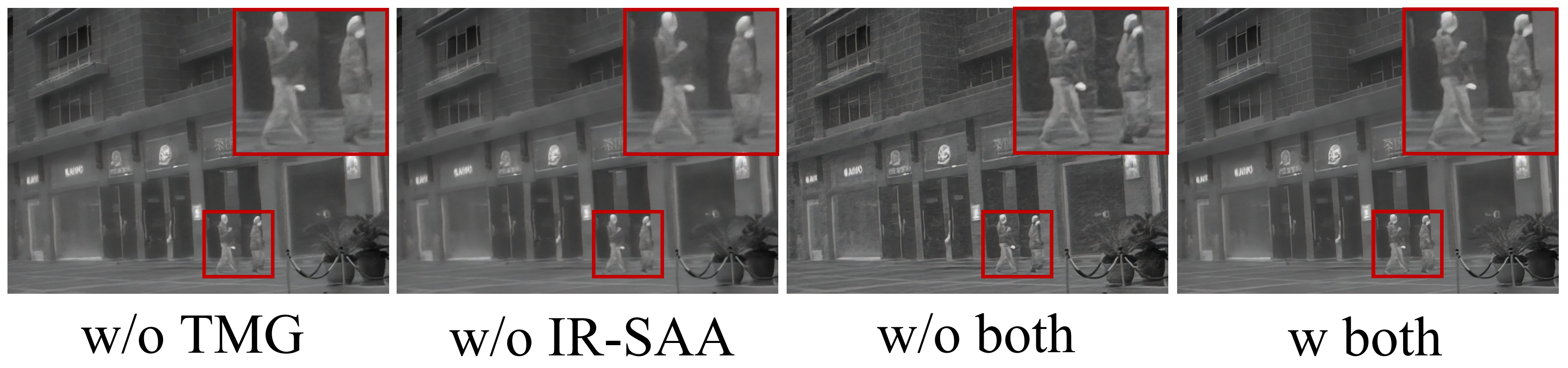}
    \caption{Qualitative ablation on the TAD.}
    \label{fig:ablation_tad}
\end{figure}

\subsection{Ablation Studies}
\subsubsection{Phasor-Guided Flow Estimator}
To validate the effectiveness of the proposed PhasorFlow, we replace it with the pre-trained optical flow network SpyNet~\cite{ranjan2017optical}. As shown in Table~\ref{tab:flow_ablation}, PhasorFlow consistently outperforms SpyNet across all evaluation metrics, with notable improvements of approximately 4.7 dB in PSNR and 12 points in VMAF. These results demonstrate that PhasorFlow leads to significant enhancements in both structural fidelity and perceptual quality of the restored videos.

We further provide qualitative comparisons as illustrated in Figure~\ref{fig:ablation_phasorflow}. Compared with the results obtained using SpyNet, PhasorFlow better preserves object boundaries, produces clearer textures, and significantly suppresses background noise. These advantages are especially evident in thermally active regions, such as human silhouettes, where PhasorFlow provides more consistent and temporally stable flow fields. The corresponding optical flow maps further intuitively highlight its ability to preserve coherent motion boundaries in these regions, while SpyNet suffers from severe distortions and fragmented flow predictions.

\subsubsection{Turbulence-Aware Decoder}
To evaluate the effectiveness of the Turbulence-Aware Decoder (TAD), we conduct an ablation study by removing its two key components: Turbulence Mask Gating (TMG) and IR Structure-Aware Attention (IR-SAA). As shown in Table~\ref{tab:ablation_tad}, the absence of either module leads to noticeable performance drops. In particular, removing IR-SAA causes a significant decline in perceptual quality. In contrast, removing TMG primarily compromises alignment robustness and fidelity, as reflected by increased LPIPS and DISTS values. The removal of both modules leads to further degradation, underscoring the necessity of multi-level turbulence modeling for reliable restoration under severe distortions.

Figure~\ref{fig:ablation_tad} presents qualitative comparisons, demonstrating that the complete TAD yields richer texture details and more coherent background structures. These results highlight the complementary contributions of TMG and IR-SAA to structural modeling and consistency preservation.

\begin{table}[!tp]
\centering
\footnotesize
\resizebox{\linewidth}{!}{%
\begin{tabular}{@{}ccccccc@{}}
\Xhline{1.1pt}
 \(M_{occ}\) & \(M_{phasor}\) & PSNR$\uparrow$ & SSIM$\uparrow$ & LPIPS$\downarrow$ & DISTS$\downarrow$ & VMAF$\uparrow$ \\
\hline
- & - & 26.3073 & 0.6558 & 0.3503 & 0.2370 & 34.0477 \\
\checkmark & - & 31.6965 & 0.7595 & 0.2866 & 0.2248 & 39.1762 \\
- & \checkmark & 28.9239 & 0.7149 & 0.2242 & 0.1817 & 30.4051 \\
\checkmark & \checkmark & \textbf{32.1595} & \textbf{0.8087} & \textbf{0.1573} & \textbf{0.1478} & \textbf{42.6042} \\
\Xhline{1.1pt}
\end{tabular}
}
\caption{Quantitative ablation on the masked guidance.}
\label{tab:ablation_heat-aware-guidance}
\end{table}

\begin{figure}[!tp]
    \centering
    \includegraphics[width=0.47\textwidth]
    {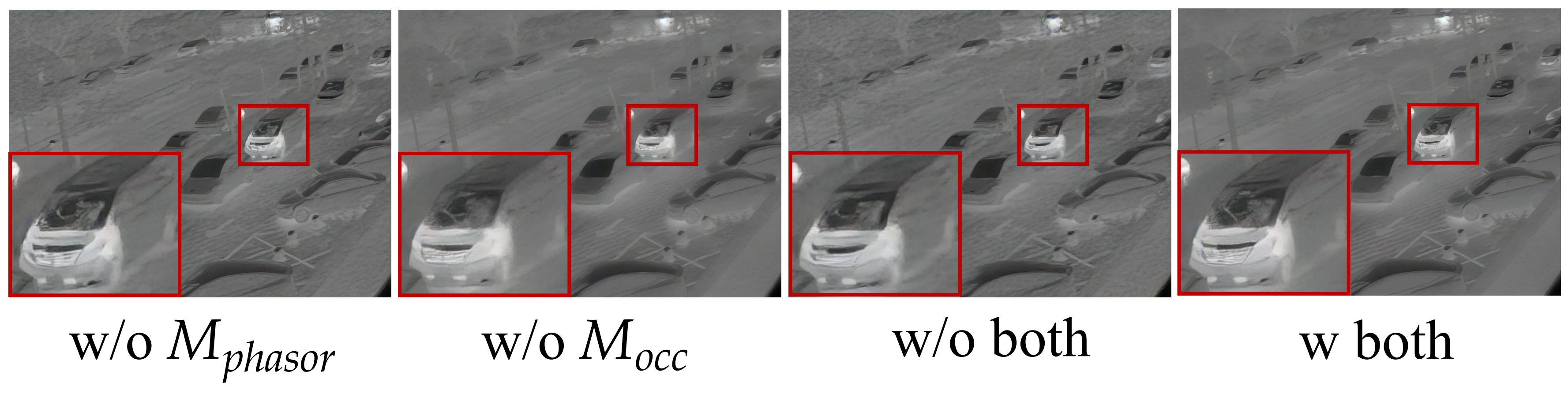}
    \caption{Qualitative ablation on the masked guidance.}
    \label{fig:ablation_heat-aware-guidance}
\end{figure}

\subsubsection{Heat-Aware Guidance}
To validate the effectiveness of the Heat-Aware Guidance mechanism, we conduct a quantitative ablation study under four configurations: (1) without any heat-aware modulation mask, (2) using only the Phasor Mask \(M_{phasor}\), (3) using only the Occlusion Mask \(M_{occ}\), and (4) applying both to form the heat-aware modulation mask \(M_{joint}\). As reported in Table~\ref{tab:ablation_heat-aware-guidance}, the joint application of both masks consistently achieves the best performance across all metrics, confirming their complementary roles in enhancing both perceptual quality and structural fidelity by localizing reliable temporal structures.

Figure~\ref{fig:ablation_heat-aware-guidance} provides qualitative evidence. Without the heat-aware modulation mask, the restored images suffer from blurred contours and structure loss, particularly in fine-grained regions such as vehicle grilles.

\section{Conclusion}
We propose \textbf{HATIR}, a heat-aware diffusion framework that unifies alignment and restoration for turbulent infrared VSR. By introducing a phasor-guided flow estimator and a turbulence-aware decoder, HATIR integrates physically grounded priors into the denoising process, enabling robust structural recovery under severe turbulence. Experiments on the newly built FLIR-IVSR dataset validate the effectiveness of our approach.

\subsubsection{Broader Impact}
HATIR enhances infrared VSR under turbulence, benefiting critical applications such as autonomous driving, surveillance, and thermal monitoring in low-visibility settings. The proposed FLIR-IVSR dataset encourages future research in infrared VSR. 
{
    \small
    \bibliographystyle{ieeenat_fullname}
    \bibliography{main}
}

\end{document}